\documentclass{ieeeaccess}
\usepackage{cite}
\usepackage{amsmath,amssymb,amsfonts}
\usepackage{booktabs}
\usepackage{algorithmic}
\usepackage{graphicx}
\usepackage{lipsum} 
\usepackage{textcomp}

\usepackage{bm}
\makeatletter
\AtBeginDocument{\DeclareMathVersion{bold}
\SetSymbolFont{operators}{bold}{T1}{times}{b}{n}
\SetSymbolFont{NewLetters}{bold}{T1}{times}{b}{it}
\SetMathAlphabet{\mathrm}{bold}{T1}{times}{b}{n}
\SetMathAlphabet{\mathit}{bold}{T1}{times}{b}{it}
\SetMathAlphabet{\mathbf}{bold}{T1}{times}{b}{n}
\SetMathAlphabet{\mathtt}{bold}{OT1}{pcr}{b}{n}
\SetSymbolFont{symbols}{bold}{OMS}{cmsy}{b}{n}
\renewcommand\boldmath{\@nomath\boldmath\mathversion{bold}}}
\makeatother

\def\BibTeX{{\rm B\kern-.05em{\sc i\kern-.025em b}\kern-.08em
    T\kern-.1667em\lower.7ex\hbox{E}\kern-.125emX}}

\begin{document}
\history{Date of publication xxxx 00, 0000, date of current version xxxx 00, 0000.}
\doi{10.1109/ACCESS.2024.0429000}

\title{Artistic Intelligence: A Diffusion-Based Framework for High-Fidelity Landscape Painting Synthesis}
\author{\uppercase{Wanggong Yang}\authorrefmark{1}, 
and Yifei Zhao\authorrefmark{1}}

\address[1]{School of New Media, Beijing Institute of Graphic Communication, Beijing, China}

\markboth
{Author \headeretal: Preparation of Papers for IEEE TRANSACTIONS and JOURNALS}
{Author \headeretal: Preparation of Papers for IEEE TRANSACTIONS and JOURNALS}

\corresp{Corresponding author: Yifei Zhao (e-mail: zhaoyifei@bigc.edu.cn).}

\tfootnote{This work was supported in part by the Beijing Municipal High-Level Faculty Development Support Program under Grant BPHR202203072.}

\begin{abstract}
Generating high-fidelity landscape paintings remains a challenging task that requires precise control over both structure and style. In this paper, we present LPGen, a novel diffusion-based model specifically designed for landscape painting generation. LPGen introduces a decoupled cross-attention mechanism that independently processes structural and stylistic features, effectively mimicking the layered approach of traditional painting techniques. Additionally, LPGen proposes a structural controller, a multi-scale encoder designed to control the layout of landscape paintings, striking a balance between aesthetics and composition.
Besides, the model is pre-trained on a curated dataset of high-resolution landscape images, categorized by distinct artistic styles, and then fine-tuned to ensure detailed and consistent output. Through extensive evaluations, LPGen demonstrates superior performance in producing paintings that are not only structurally accurate but also stylistically coherent, surpassing current state-of-the-art models. This work advances AI-generated art and offers new avenues for exploring the intersection of technology and traditional artistic practices. Our code, dataset, and model weights will be publicly available.
\end{abstract}


\titlepgskip=-21pt

\maketitle

\section{Introduction}
\label{sec:introduction}
\PARstart{L}{andscape} painting holds a significant place in traditional art, depicting natural scenes to convey the artist's aesthetic vision and emotional connection to nature~\cite{kang2020compressed, wang2023attentionGAN, kang2021tunable, lin2021styleTransfer, lu2022tileImagesGAN}. This genre utilizes tools such as ink, water, paper, and brushes, emphasizing artistic conception, brush and ink intensity variations, balanced composition, and effective use of negative space~\cite{yang2022deep, sun2023paint}. With the development of computer technology, especially diffusion, the generation of landscape paintings through modern algorithms has become a promising area of research~\cite{sun2023paint}.

\begin{figure}[t]
\centering
\includegraphics[width=1\linewidth]{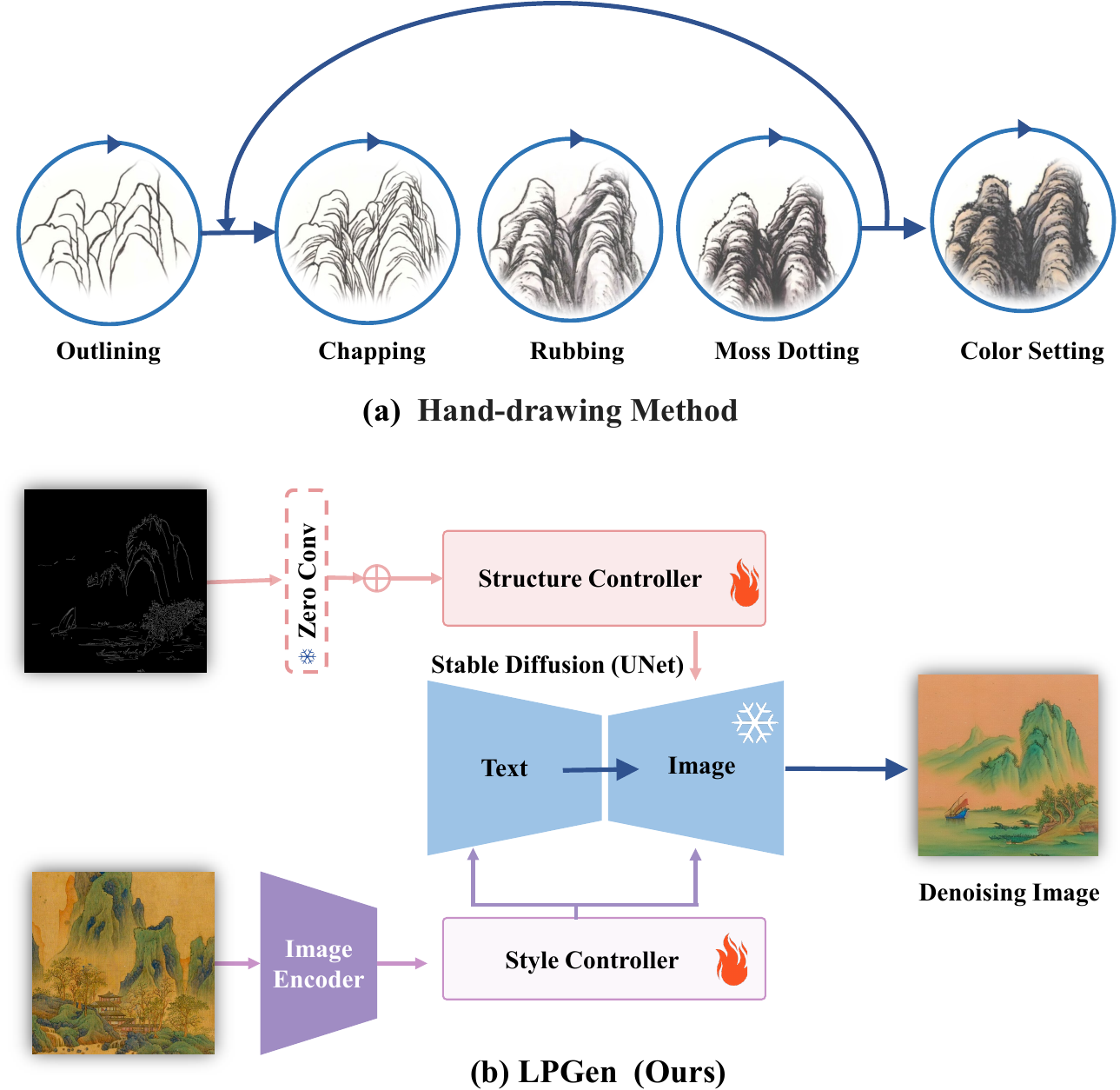}
\caption{Hand-drawing method compared with the proposed LPGen method. The hand-drawing method involves a complex and repetitive process with several key steps such as outlining, chapping, rubbing, moss-dotting, and coloring. LPGen simplifies the process, making it convenient and flexible by precisely generating landscape paintings from a specified style reference image and a canny outline image.}
\label{fig:fig1_Processes}
\end{figure}

The creative process of landscape painting involves several intricate stages: outlining, chapping, rubbing, moss-dotting, and coloring, as shown in Figure~\ref{fig:fig1_Processes} (a)~\cite{sun2023paint}. Outlining is the initial stage where the artist sketches the basic structure and composition of the landscape~\cite{sun2023paint, yao2024enhancing}. Chapping involves the careful depiction of textures, requiring an understanding of how to use brushstrokes to convey different natural surfaces. Rubbing and dyeing necessitate meticulous control over ink intensity and moisture, balancing between depth and subtlety in shading~\cite{sun2023paint}. Moss-dotting requires a flexible and natural technique to create the illusion of texture and dimension, adding complexity and life to the painting. Finally, coloring demands a keen sense of color harmony to ensure the overall unity and aesthetic appeal of the artwork~\cite{yao2024enhancing}. The complexity and precision involved in each stage contribute to the overall challenge of creating landscape paintings~\cite{sun2023paint, yao2024enhancing}.

Recent technological advancements, particularly in deep learning, have empowered computers to emulate landscape paintings' artistic styles and brushwork. These technologies can generate artworks that capture the rich cultural essence of China, preserving traditional art and opening new paths for creative expression~\cite{yang2022deep, li2024deep, way2023structure, sun2023dataset}. This technological innovation supports the preservation and evolution of conventional culture~\cite{way2023structure, sun2023dataset, way2023twingan}.
Research into the generation of landscape paintings is primarily divided into traditional and deep learning-based methods. Traditional methods include non-photorealistic rendering, image-based approaches, and computer-generated simulations. Non-photorealistic rendering uses computer graphics to simulate various artistic effects, such as ink diffusion, through specialized algorithms~\cite{zhang2022ink, semmo2022comprehensive}. Image-based methods involve collecting brush stroke texture primitives (BSTP) from hand-drawn samples and employing them to map multiple layers to create mountain imagery~\cite{kim2024dals, lu2023interactive}. Despite their sophistication, the quality of computer-generated simulations often falls short~\cite{zhang2022ink}.

Currently, many researchers are using generative adversarial networks (GANs) to generate landscape paintings~\cite{peng2022contour, xue2021landscapeGAN, ma2023styleTransfer}. However, due to the distinctive artistic style of landscape painting, GAN-based methods often face challenges such as unclear outlines, uncontrollable details, and inconsistent style migration~\cite{guo2022aiArtGAN}.
By visual generation, such as stable diffusion (SD), the artistic effects of landscape painting can be effectively simulated and reproduced~\cite{rombach2022ldm, wang2023stableDiffusion}. Diffusion models (DMs) have been successfully applied across various domains, including clothing swapping, face generation, dancing, and anime creation, demonstrating their versatility in generating high-quality realistic content from diverse inputs~\cite{ho2020ddpm, li2024uvIDM, joby2024synthesizing}. For instance, diffusion models can produce highly realistic images nearly indistinguishable from actual photographs in face generation~\cite{peng2023sketch, kim2023dcface}. Similarly, in anime creation, they can generate detailed and stylistically consistent images from simple prompts~\cite{jeon2024cartoonizediff}. Although this method preserves the stylistic features of traditional landscape paintings, it does not allow precise control over the structure and style~\cite{wang2023stableDiffusion, rombach2022ldm}.

Structure and style control are crucial factors in landscape painting, significantly influencing the overall effect and quality of the artwork. Previous GAN-based and Diffusion-Based methods have often yielded unsatisfactory results in generating landscape paintings. To address this, we propose the LPGen framework, which draws inspiration from the fundamental steps of traditional landscape painting. The framework consists of two parts: structure controller and style controller. The framework uses canny line graphs to control the painting structure through the structure controller and uses the reference style graph to control the style of the generated painting through the style controller. The main contributions of this work are summarized as follows:
\begin{itemize}
\item[\textcolor{black}{$\bullet$}]We propose LPGen, a high-fidelity, controllable model for landscape painting generation. This model introduces a novel multi-modal framework by incorporating image prompts into the diffusion model.
\item[\textcolor{black}{$\bullet$}]We construct a comprehensive dataset comprising 2,416 high-resolution images, meticulously categorized into three distinct styles: azure green landscape, ink wash landscape, and light vermilion landscape.
\item[\textcolor{black}{$\bullet$}]
We conduct extensive qualitative and quantitative analyses of our proposed model, LPGen, providing clear evidence of its superior performance compared to several state-of-the-art methods.
\end{itemize}

\section{Related Work}
\subsection{GAN-Based Method} 
Due to the rapid advancements in deep learning, significant progress has been made in various visual tasks~\cite{hayoun2024physics, zhang2022beyond}. Based on the input content, landscape painting generation methods can be categorized into three types: nothing-to-image, image-to-image, and text-to-image. For the first type, it does not require any input information to generate landscape paintings. An example is the Sketch-And-Paint GAN (SAPGAN)~\cite{xue2021end}, the first neural network model capable of automatically generating traditional landscape paintings from start to finish. SAPGAN consists of two GANs: SketchGAN for generating edge maps and PaintGAN for converting these edge maps into paintings. Another example is an automated creation system~\cite{luo2022high} based on GANs for landscape paintings, which comprises three cascading modules: generation, scaling, and super-resolution.

For the second type, a photo is input for style transfer. An interactive generation method~\cite{zhou2019interactive} allows users to create landscape paintings by simply outlining with essential lines, which are then processed through a recurring adversarial generative network (RGAN) model to generate the final image. Another approach, neural abstraction style transfer~\cite{li2019neural}, leverages the MXDoG filter and three fully differentiable loss terms. The ChipGAN architecture~\cite{sun2022style} is an end-to-end GAN model that addresses critical techniques in ink painting, such as blanks, brush strokes, ink tone and spread. Furthermore, an attentional wavelet network~\cite{sun2023paint} utilizes wavelets to capture high-level mood and local details of paintings via the 2-D Haar wavelet transform. Lastly, the Paint-CUT model~\cite{wang2023cclap} intelligently creates landscape paintings by utilizing a shuffle attentional residual block and edge enhancement techniques.

For the third type, the output image is guided by textual input information. A novel system~\cite{he2023composing} transforms classical poetry into corresponding artistic landscape paintings and calligraphic works. Another approach, controlled landscape painting generation (CCLAP)~\cite{wang2023cclap}, is based on the latent diffusion model (LDM)~\cite{rombach2022high} and comprises two sequential modules: a content generator and a style aggregator. Although this method allows text-based control of the image, it still faces challenges related to the randomness of image generation and limited controllability.

\subsection{Diffusion-Based Method}  
DM~\cite{ho2020denoising} is a deep learning framework primarily utilized for image processing and computer vision tasks. Fundamentally, DM work by corrupting the training data by continuously adding Gaussian noise, and then learning to recover the data by reversing this noise process. DM can generate detailed images from textual descriptions and are also effective for image restoration, image drawing, text-to-image, and image-to-image tasks~\cite{lin2023diffbir, zhang2024mrir, zhang2024forget, fei2023generative}. Essentially, by providing a textual description of the desired image, SD can produce a realistic image that aligns with the given description~\cite{ho2020denoising}. These models reframe the "image generation" process into a "diffusion" process that incrementally removes noise. Starting with random Gaussian noise, the models progressively eliminate it through training until it is entirely removed, yielding an image that closely matches the text description~\cite{ho2020denoising, zhang2024mrir}. However, a significant drawback of this approach is the considerable time and memory required, especially for high-resolution image generation~\cite{lin2023diffbir}. LDM~\cite{rombach2022high} was developed to address these limitations by significantly reducing memory and computational costs. This is achieved by applying the diffusion process within a lower-dimensional latent space instead of the high-dimensional pixel space~\cite{lin2023diffbir}.

The emergence of large text-to-image models that can create visually appealing images from brief descriptive prompts has highlighted the remarkable potential of AI. However, these models encounter challenges such as uneven data availability across specific domains compared to the generalized image-to-text domain. In addition, some tasks require more precise control and guidance than simple prompts can provide. ControlNet~\cite{zhang2023adding} addresses these issues by generating high-quality images based on user-provided cues and controls, which can fine-tune performance for specific tasks. Meanwhile, IP-Adapter~\cite{ye2023ip} employs decoupled cross-attention mechanisms for the characteristics of the text and the image. Inspired by these approaches, our study incorporates ControlNet's additional structural controls and the IP-Adapter's style guiding the generation of landscape paintings.


\section{Proposed Method}
To address the above issues, LPGen has been developed to generate landscape paintings from canny image and structure image. This section provides an overview of the framework, the structure controller, which manages the structure of the generated image, and the style controller, which dictates the painting style of the resulting image.

\begin{figure*}[t]
\centering
\includegraphics[width=1\linewidth]{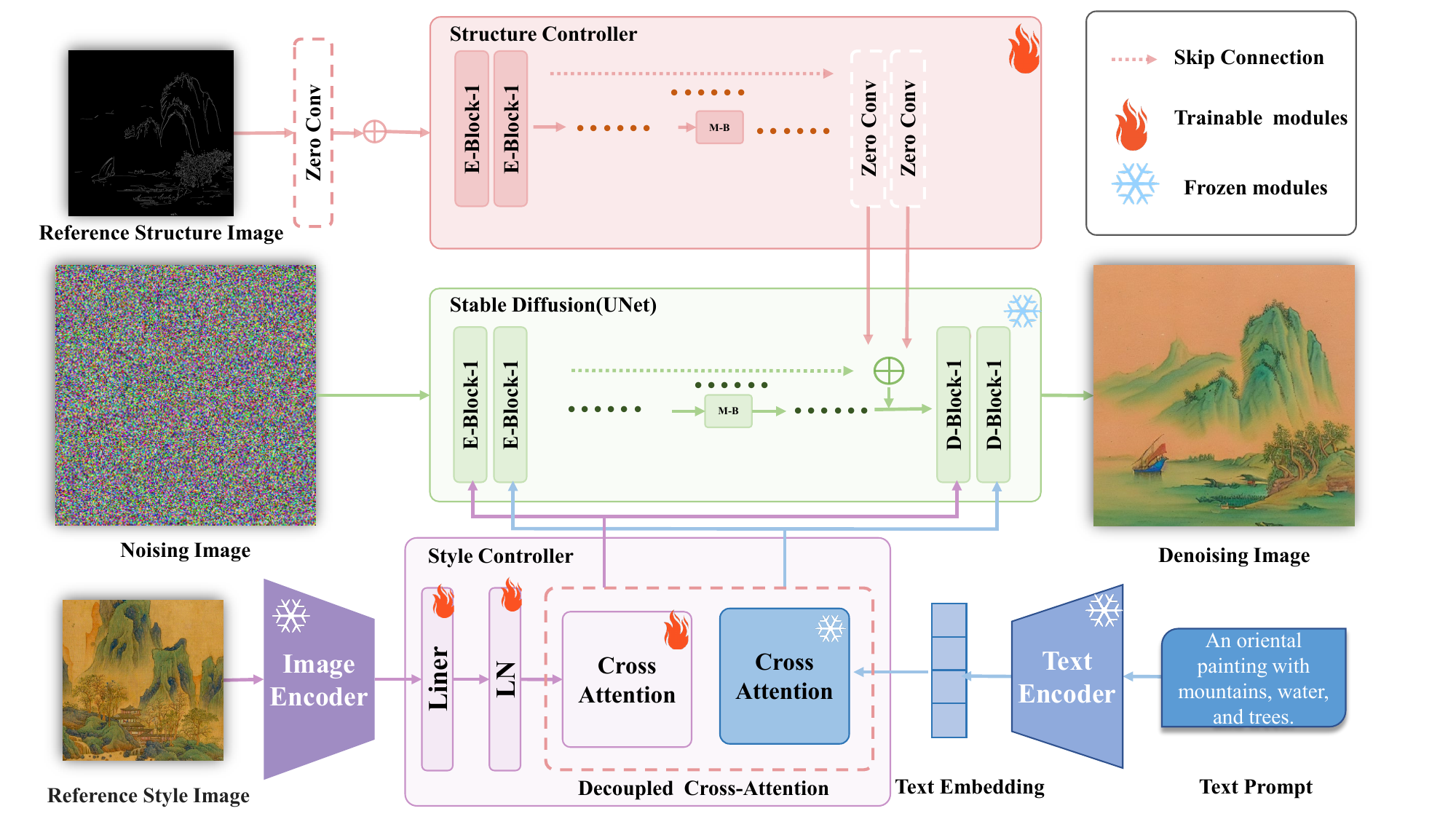}
\caption{Schematic of LPGen for landscape painting generation
The schematic comprises two key components: the structure controller and the style controller. The structure controller module utilizes the Decoupled Cross-Attention technique to separately manage the structural information of an image across different domains, allowing for precise control and regulation of specific elements and attributes in the generated image. The style controller dynamically adjusts the features of the input image, enabling the generative model to accurately capture and reflect the style and structure of the source image.}
\label{fig:fig2_SystemArchitecture}
\end{figure*}

\subsection{Preliminaries}
The complete structure of the LPGen framework is illustrated in Figure~\ref{fig:fig2_SystemArchitecture}. The framework comprises three main components: the stable diffusion model, the structure controller, and the style controller. 

While the stable diffusion model offers generalized image generation capabilities, it lacks precise control over the structural and stylistic features of an image. To address the structural control problem, we use a structural controller to ensure edge information and effectively guide the image generation process. The structural controller determines the position and shape of graphic elements by outlining, so that the generated image is aligned with the structure of the refined reference image. For stylistic aspects, the style controller extracts the style of a reference segmented image and applies it to the generated image, allowing the generation of images in a specified style. The style controller learns various color schemes and brush features to convey different emotions. 

Building upon the stable diffusion model’s basic image generation capabilities, the structure controller manages the structural layout, and the style controller governs the color scheme features. This precise and robust control enables us to achieve excellent management over the style and structural of the generated images, surpassing the capabilities of simple text-to-image models.

\subsection{Structure Controller} 
The outlining serves as the fundamental framework of landscape painting, establishing the overall layout and positioning of elements. To enable the canny like as outlining of  Hand-drawing Method to guide LDM in generating the image, the structure controller manipulates the neural network structure of the diffusion model by incorporating additional conditions. This, in combination with Stable Diffusion, ensures accurate spatial control, effectively addressing the spatial consistency issue. The image generation process thus emulates the actual painting process, transitioning from outlining to coloring. We employ the structure controller as a model capable of adding spatial control to a pre-trained diffusion model beyond basic textual prompts. This controller integrates the UNet architecture from Stable Diffusion with a trainable UNet replica. This replica includes zero convolution layers within the encoder and middle blocks. The complete process executed by the structure controller is as follows:

\begin{equation}
    y_{c} = \mathcal{F}(x,\theta)+ \mathcal{Z}(\mathcal{F}(x++ \mathcal{Z}(c,\theta_{z1}),\theta_{c}),\theta_{z2}).
\end{equation}

The structure controller differentiates itself from the original SD in handling the residual component. $\mathcal{F}$ signifies the UNet architecture, with $x$ as the latent variable. The fixed weights of the pre-trained model are denoted by $\theta$. Zero convolutions are represented by $\mathcal{Z}$, having weights $\theta_{z1}$ and $\theta_{z2}$, while $\theta_{c}$ indicates the trainable weights unique to the structure controller. Essentially, the structure controller encodes spatial condition information, such as that from canny edge detection, by incorporating residuals into the UNet block and subsequently embedding this information into the original network.

\subsection{Style Controller} 
During the coloring stage, the artist sequentially fills and renders the picture, coloring each element according to the guidelines established by the outlines. In the original SD base model with text-to-image generation capability, we introduce the style controller structure to integrate prompt features, structure features, and style features. In order to achieve the above functions, our novel design is a decoupled cross-attention mechanism as shown in Figure~\ref{fig:fig2_SystemArchitecture} , where image features are embedded through a newly added cross-attention layer, which consists mainly of two modules: an image encoder for extracting image features, and an adaptation module with decoupled cross-attention for embedding the image features into a pre-trained text-to-image diffusion model.

The pre-trained CLIP image encoder model is used, but here, in order to efficiently decompose the global image Embedding, we use a small trainable projection network to project the image Embedding into a sequence of features of length N=4. The projection network here is designed as a linear layer Linear plus a normalization layer LN, and at the same time, the dimensions of the input image features are kept consistent with the dimensions of the text features in the pre-trained diffusion model.

\begin{figure*}[t]
\centering
\includegraphics[width=1\linewidth]{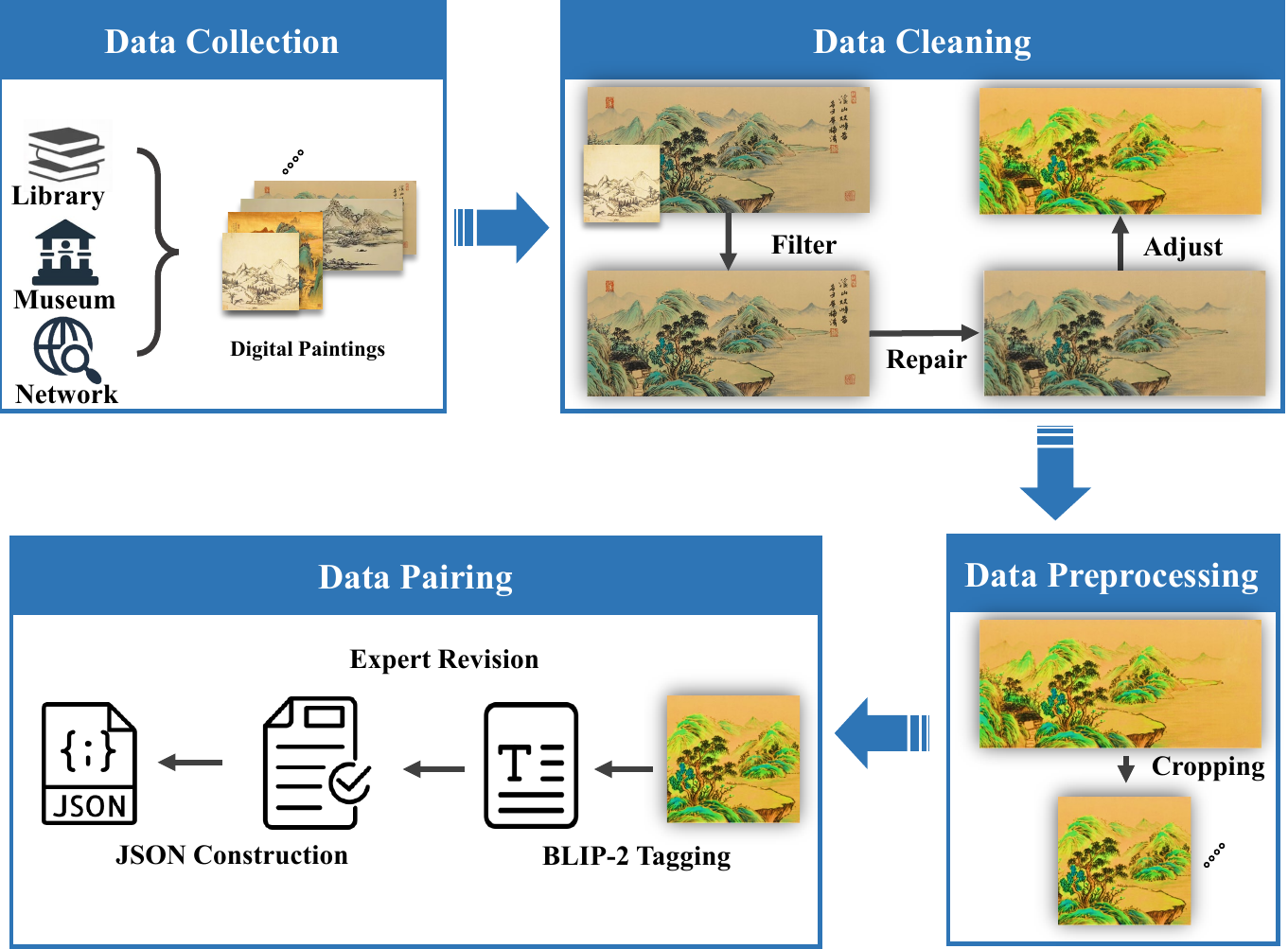}
\caption{Experimental dataset processing workflow. The workflow begins with collecting raw image data, followed by cleaning the data to eliminate noise, duplicates, and irrelevant information. The data is then pre-processed by resizing images and converting formats. Finally, matching pairs of text and images are created for model training.}
\label{fig:fig3_DatasetWorkflow}
\end{figure*}

In the original Stable Diffusion model, text embeddings are injected into the Unet model through the input-to-cross-attention mechanism. A straightforward way to inject image features into the Unet model is to join image features with text features and then feed them together into the cross-attention layer. However, this approach is not effective enough. Instead, the style controller separates the cross-attention layer for text features and image features by decoupling this mechanism, making the model more concise and efficient. This design not only reduces the demand for computational resources, but also improves the generality and scalability of the model. During the training process, the style controller is able to automatically learn how to generate corresponding images based on text descriptions, while maintaining the effective utilization of image features. This enables style controller to generate images with full consideration of the semantic information of the text, thus generating more accurate and realistic images.

The text corresponds to the cross-attention as:

\begin{equation}
    \mathcal{Z}_{new} = \mathcal{\textit{Attention}}(Q,K^{t},V^{t}) + \lambda \cdot \mathcal{\textit{Attention}}(Q,K^{i},V^{i}).
\end{equation}

Here, $Q$, $K^{t}$, and $V^{t}$ represent the query, key, and value matrices for the text cross-attention operation, while $K^{i}$ and $V^{i}$ are the key and value matrices for the image cross-attention. Given the query features $Z$ and the image features $c_{i}$, the formulations are as follows: $Q = ZW_{q}$, $K^{i} = c_{i}W_{k}^{i}$, and $V^{i} = c_{i}W_{v}^{i}$. It is important to note that only $W_{k}^{i}$ and $W_{v}^{i}$ are trainable weights.
\subsection{Training and Inference} 
During training, we focus solely on optimizing the style controller, leaving the parameters of the pre-trained diffusion model unchanged. The style controller is trained using a dataset of paired images and text. Still, it can train without text prompts, as only image prompts effectively guides the final generation. The training objective remains consistent with that of the original SD:
\begin{equation}
L_{\text{simple}}=\mathbb{E}_{\boldsymbol{x}_{0},\boldsymbol{\epsilon}, \boldsymbol{c}_{t}, \boldsymbol{c}_{i}, t} \| \boldsymbol{\epsilon}- \boldsymbol{\epsilon}_\theta\big(\boldsymbol{x}_t, \boldsymbol{c}_{t}, \boldsymbol{c}_{i}, t\big)\|^2.
\end{equation}

During the training stage, we consistently employ the random omission of image conditions to facilitate classifier-free guidance during inference. If the image condition is omitted, we replace the CLIP image embedding with a zero vector.
\begin{equation}
\begin{split}
\hat{\boldsymbol{\epsilon}}_{\theta}(\boldsymbol{x}_t, \boldsymbol{c}_{t}, \boldsymbol{c}_{i}, t) = w\boldsymbol{\epsilon}_{\theta}(\boldsymbol{x}_t, \boldsymbol{c}_{t}, \boldsymbol{c}_{i}, t)+(1-w)\boldsymbol{\epsilon}_{\theta}(\boldsymbol{x}_t, t).
\end{split}
\end{equation}

Since text cross-attention and image cross-attention are separate, we can independently adjust the weight of the image condition during inference.

\begin{equation}
\mathbf{Z}^{new}=\text{Attention}(\mathbf{Q},\mathbf{K},\mathbf{V}) + \lambda\cdot\text{Attention}(\mathbf{Q},\mathbf{K}',\mathbf{V}'),
\end{equation}
here, $\lambda$ is a weighting factor. When $\lambda=0$, the model defaults to the original text-to-image diffusion model.

\subsection{Datasets}

Text-to-image generative diffusion models have been slow to develop in landscape painting generation due to the lack of image-text paired datasets describing style and content. Thus, the diffusion model cannot stably generate landscape painting
for specified decoration style and structure. To this end, it is imminent to establish a new dataset of interior design decorating styles. To this end, this study first constructed a dataset, LPD-3, with descriptions of styles and content, which were collected by from websites, as illustrated in Figure~\ref{fig:fig3_DatasetWorkflow}. We have expanded our collection to include various styles of landscape painting and standardized the corresponding text descriptions. This effort aims to contribute positively to the research on landscape painting generation.

\emph{Data Collection.}
Datasets serve as the foundation for the rapid advancement of artificial intelligence, with a few open datasets significantly contributing to this progress. For example, the traditional landscape painting dataset ~\cite{xue2021end} is the only accessible dataset, but it has limitations in terms of data volume and lacks categorization or identification of the image data. To address this problem, we firstly have curated a collection of digital paintings sourced through websites such as Baidu, artwork websites, photographs from landscape painting books, and digital museum databases. In the following, we enlisted the help of professors and landscape artists to assist in the classification process. All the landscape paintings are classified into three main categories: azure green landscape, ink wash landscape, and light vermilion landscape. 

\emph{Data Cleaning.}
 To ensure the quality of the dataset, these experts manually removed paintings that did not depict landscapes and those with dimensions smaller than 512 pixels or with unclear imagery. If the collected images are used directly for training, the resulting landscape paintings may contain inexplicable elements or words. To better emphasize the theme and beauty of landscape paintings while removing the interference of calligraphy and seal cutting, we utilized image processing software like Adobe Photoshop to repair the damaged area, enhancing image clarity and quality. We adjusted the brightness, contrast, hue, and saturation parameters to make the images sharper and brighter and to highlight their details and features.

\emph{Data Preprocessing.}
Since the training data for this study should be in $512 \times 512$ format, preprocessing the cleaned data is an essential step. First, each image was scaled so that its shorter side was 512 pixels, maintaining the original aspect ratio. For images with an aspect ratio less than 1.5, a $512 \times 512$ section was cropped from the center and saved. For images with an aspect ratio greater than 1.5, after cropping the $512 \times 512$ section from the center, the center point was moved 256 pixels in both directions along the longer side to crop and save two additional images. If the cropped image was not square, the process was halted, and the current cropped image was discarded. The distribution of the number of pictures is shown in Table~\ref{tab:image_counts}.

\begin{table*}[h]
\centering
    \caption{Landscape painting type distribution of three distinct styles in our collected dataset. The data comes from Harvard, Smithsonian, Metropolitan, Baidu, and Princeton.}
    \label{tab:image_counts}
    \hspace{-1cm}
    \begin{tabular}{l| c c c c c}
        \toprule
        \emph{} & \emph{Harvard}  & \emph{Smithsonian} & \emph{Metropolitan} & \emph{Baidu} & \emph{Princeton}\\
        \midrule
        Azure Green Landscape & 12   & 192  & 19   & 181  & 18    \\
        Light Vermilion Landscape & 18 & 742 & 256 & 104 & 273 \\
        Ink Wash Landscape & 68 & 342 & 119 & 20 & 52  \\
        \hline
        Total & 98 & 1276 & 394 & 305 & 343  \\
        \bottomrule
    \end{tabular}
\end{table*}

\emph{Data Pairing.}
The data format for this experiment is image-text pairs because it allows the model to learn the association between visual content and descriptive language, enhancing its ability to generate images that align with specific textual descriptions, thereby improving the overall accuracy and relevance of the generated images. Although image-text pairs previously required manual annotation, we now utilize BLIP-2 to automatically generate the corresponding text information for the collected images. Although the text generated by BLIP-2 ~\cite{li2022blip} may be inaccurate, we invited artists who curated descriptive texts specific to landscape paintings. To increase the diversity of text descriptions, we provided multiple expressions for each text and specified the painting type within the descriptions. Finally, we saved the image paths and corresponding text descriptions in JSON files. 

\subsection{Evaluation Metrics}
Several metrics are commonly used to assess how closely generated images resemble reference images in evaluating image and structural similarity. This document explains the meanings, formulas and applications of six key metrics: the learned perceptual image patch similarity (LPIPS)~\cite{zhang2018perceptual}, gram matrix~\cite{gatys2015texture}, histogram similarity~\cite{swain1991color}, chamfer match score~\cite{barrow1977parametric}, hausdorff distance~\cite{huttenlocher1993comparing}, and contour match score~\cite{belongie2002shape}.

\emph{Learned Perceptual Image Patch Similarity.} LPIPS is a metric to evaluate image similarity. It utilizes feature representations of deep learning models to measure the perceptual similarity between two images. Unlike traditional pixel-based similarity metrics, LPIPS focuses more on the perceptual quality of the image, that is, the perception of image similarity by the human visual system.
The formula for LPIPS is given by:
\begin{equation}
\text{LPIPS}(I_1, I_2) = \sum_{l} \frac{1}{H_l W_l} \sum_{h, w} \| w_l \cdot (\phi_l(I_1)_{h,w} - \phi_l(I_2)_{h,w}) \|_2^2,
\label{eq:lpips}
\end{equation}
where:
\( \phi_l \) denotes the feature representation at layer \( l \).
\( H_l \) and \( W_l \) are the height and width of the feature map on layer \( l \).
\( w_l \) is a learned weight at layer \( l \).
\( I_1 \) and \( I_2 \) are the two compared images.

\emph{Gram Matrix.} The gram matrix is a metric for evaluating the stylistic similarity of images and is commonly used in style migration tasks. It captures the texture information and stylistic features of an image by computing the inner product between the feature maps of a convolutional neural network. Specifically, the gram matrix describes the correlation between different channels in the feature map, thus reflecting the overall texture structure of the image.

The formula for the gram matrix at a particular layer is:
\begin{equation}
G_{ij}^l = \sum_k F_{ik}^l F_{jk}^l,
\end{equation}
where:
\( F_{ik}^l \) is the activation of the \( i \)-th channel at layer \( l \).
The sum \( \sum_k \) is taken over all spatial locations of the feature map.

\emph{Histogram Similarity (Bhattacharyya Distance).} The histogram similarity is a metric for evaluating the similarity of color distributions of two images and is widely used in image processing and computer vision. By comparing the color histograms of the images, the similarity of the images in terms of color can be determined.The bhattacharyya Distance ~\cite{bhattacharyya1943measure} is a commonly used histogram similarity metric, especially for similarity calculation of probability distributions.

The formula for the bhattacharyya distance is:
\begin{equation}
D_B(p, q) = -\ln \left( \sum_{x \in X} \sqrt{p(x) q(x)} \right),
\end{equation}
where:
Let \( p(x) \) and \( q(x) \) represent the probability distributions of the two histograms to be compared. The summation \( \sum_{x \in X} \) is taken in all the bins of the histograms.

\emph{Chamfer Match Score.} The chamfer match score is a metric for evaluating the similarity of two sets of point clouds, widely used in computer vision and graphics. It quantifies the shape similarity between two sets of point clouds by calculating the average nearest distance between them. Specifically, chamfer match score is a variant of Chamfer Distance and is commonly used for tasks such as 3D shape matching, image alignment and contour alignment.

The formula for the chamfer distance is:
\begin{equation}
d_{\text{Chamfer}}(A, B) = \frac{1}{|A|} \sum_{a \in A} \min_{b \in B} \|a - b\| + \frac{1}{|B|} \sum_{b \in B} \min_{a \in A} \|a - b\|,
\end{equation}
where:
\(A\) and \(B\) represent the sets of edge points in two images.
\(\|a - b\|\) denotes the Euclidean distance between points \(a\) and \(b\).

\emph{Hausdorff Distance.} The hausdorff distance is a metric for evaluating the maximum distance between two sets of point sets (point clouds, contours, etc.) and is widely used in computer vision, graphics, and pattern recognition. It measures the distance between the farthest corresponding points in two point sets, reflecting the degree to which they differ geometrically.

The formula for the hausdorff distance is:
\begin{equation}
d_H(A, B) = \max \left\{ \sup_{a \in A} \inf_{b \in B} \|a - b\|, \sup_{b \in B} \inf_{a \in A} \|a - b\| \right\},
\end{equation}
where:
\(A\) and \(B\) denote the sets of edge points in the two images.
\(\|a - b\|\) represents the Euclidean distance between the points \(a\) and \(b\).
\(\sup\) signifies the supremum (the least upper bound) and \(\inf\) indicates the infimum (the highest lower bound).

\emph{Contour Match Score.} The contour match score evaluates the similarity between the contour shapes of two images, making it particularly useful in applications where the overall shape and structure of objects are crucial. This score is determined by comparing the shape descriptors of the contours in the images.

The formula for the contour match score is:
\begin{equation}
d_{\text{Contour}}(A, B) = \sum_{i=1}^n \left( \frac{(A_i - B_i)^2}{A_i + B_i} \right),
\end{equation}
where:
\(A_i\) and \(B_i\) represent the shape descriptors of the contours in the two images.
The summation is taken over all contour points \(i\).

\section{Experiment and Analysis}

\subsection{Generative Controllability}

In order to prove that our proposed method can accurately control the structure and style of generated landscape images, we use different canny and style reference images as input to the LPGen model to test the effect of its generated images.

\begin{figure}
\centering
\includegraphics[width=1\linewidth]{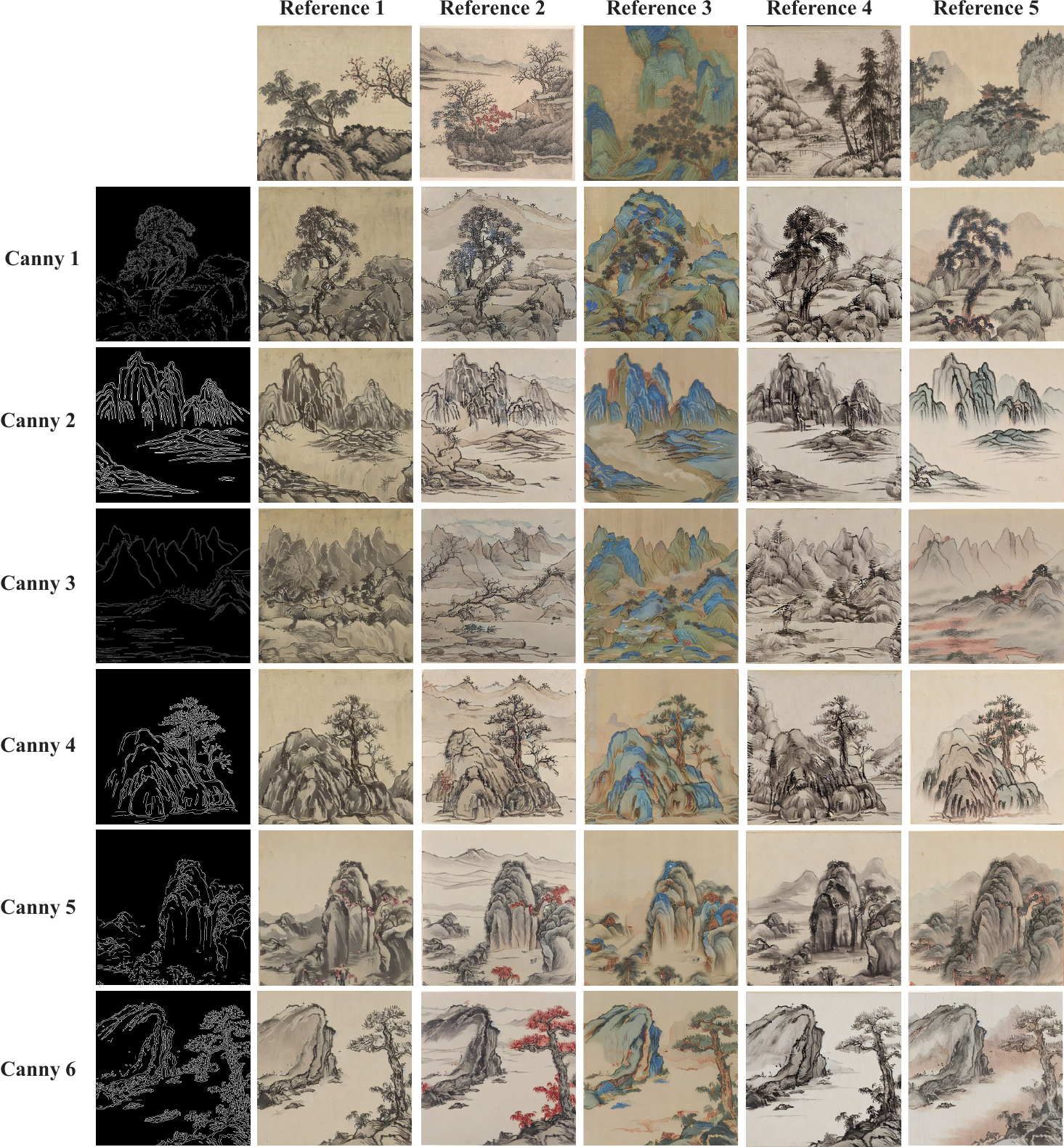}
\caption{Diverse landscape paintings generated by our method. Our method, LPGen, is  capable of producing artworks in various styles and creating differentiated paintings within the same style. Reference 1 through Reference 6 represent landscape paintings generated in the same style but with different canny edge maps. Canny 1 through Canny 6 depict landscape paintings generated with the same Canny edge map but using different style references.}
\label{fig:fig4_ExamplesbyLPen}
\end{figure}

\begin{figure*}[t]
\centering
\includegraphics[width=1\linewidth]{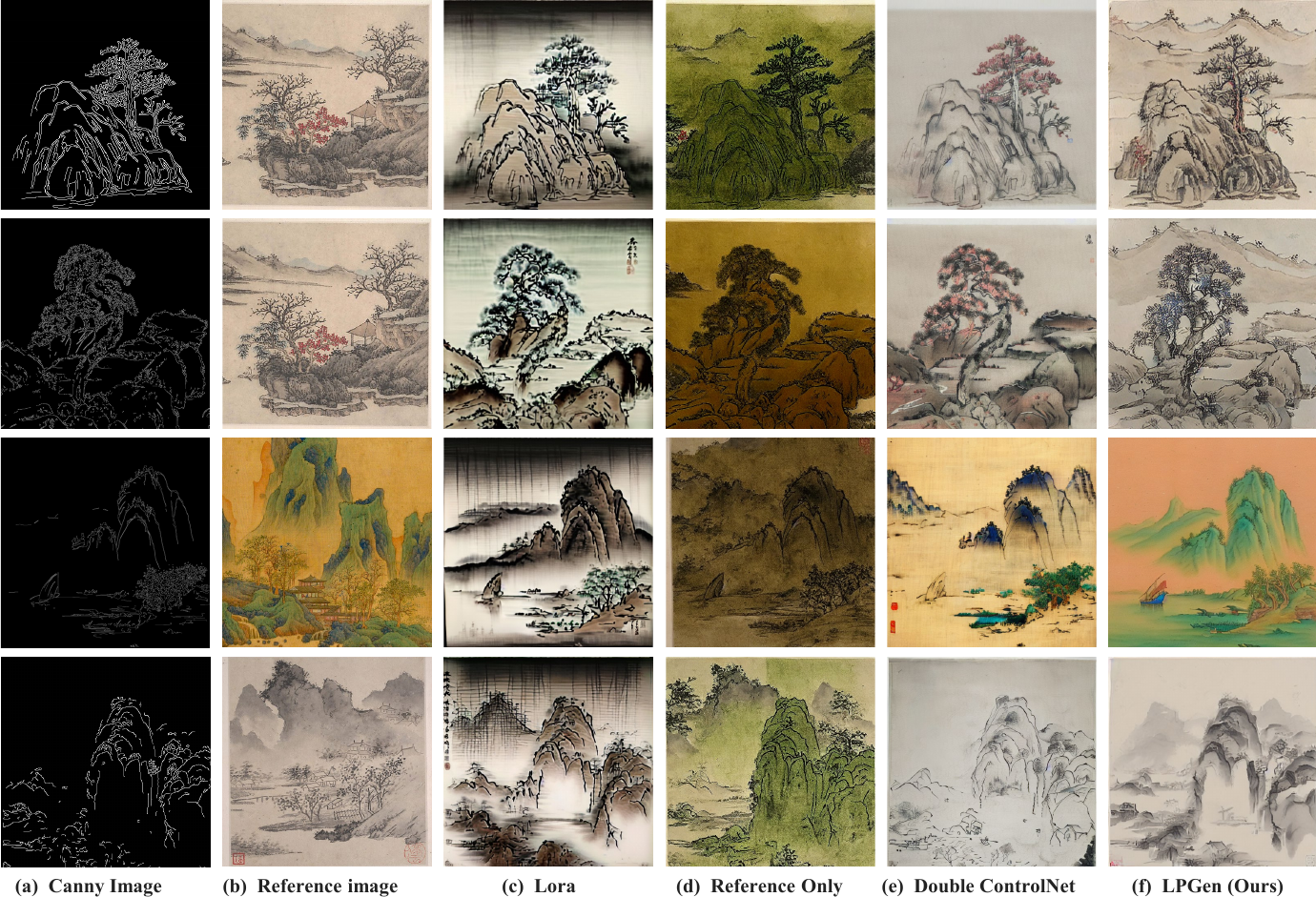}
\caption{Comparison between the proposed method and mainstream methods in generative landscape paintings. Figure 6a displays the constraint canny edge map, while Figure 6b shows the target ink-style reference image. The proposed method uses a Canny image to control the structure and a reference image to control the generated painting style. Each method generated landscape paintings using four different Canny edge maps, resulting in a total of 16 images.}
\label{fig:fig5_ComparisonResults}
\end{figure*} 

As shown in Figure~\ref{fig:fig4_ExamplesbyLPen}, we use the LPGen model to generate landscape paintings. The images in each column are landscapes generated by our proposed method in the same canny and different reference images. It is obvious that they can maintain the same structure and style as the reference image. For example, the reference image in Reference 3 is a azure green landscape style image. With different canny inputs, images with different structures are obtained. Obviously, its azure green landscape style is maintained. These experiments indicated that LPGen can learn the characteristics and styles of numerous classic landscape paintings and automatically create new highly realistic images. The generated landscape paintings feature typical natural elements such as mountains, rivers, trees, and stones and retain the ink color and brush techniques characteristic of traditional landscape paintings. By combining these two control methods, the LPGen model can generate highly realistic landscape paintings and flexibly adjust the style and details of the images, achieving precise control and innovative results in traditional art creation.


\subsection{Qualitative Analysis}

To accurately and objectively assess the effectiveness of the proposed LPGen model in generating high-quality landscape paintings, we conduct a comprehensive comparative analysis against several current state-of-the-art methods, including Reference Only \cite{zhang2023adding}, Double ControlNet ~\cite{zhang2023adding}, and Lora ~\cite{hu2021lora}.

Figure~\ref{fig:fig5_ComparisonResults} (c) is generated by the Lora method. It has obvious noise and erroneous image information and cannot revert the style of the reference image. Compared with the Lora method, the image generated by the Reference Only method has a more accurate contour structure but introduces redundant contours, with indistinct style features and inaccurate colors, as shown in Figure~\ref{fig:fig5_ComparisonResults} (d). The advantage of the Double ControlNet method is that the generated landscape paiting contain clear  structure , but cannot effectively learn the style features of the reference image, as shown in Figure~\ref{fig:fig5_ComparisonResults} (e).

Compared to other methods, the pros and cons of these methods are shown in Figure~\ref{fig:fig5_ComparisonResults}, which shows that the method proposed in this study was the best among all tested  state-of-the-art methods. The proposed LPGen effectively addresses significant issues such as poor style transfer and blurred lines in generated images, resulting in high-quality landscape paintings. LPGen not only preserves the style and structure of the target photos but also successfully captures the essence of the ink wash style, thereby achieving superior overall quality and fidelity.

\subsection{Quantitative analysis}
\begin{table*}[h]
\centering
\caption{Quantitative comparison of the proposed LPGen with several state-of-the-art models. LPIPS evaluates perceptual similarity, gram matrix (GM) measures texture correlations for style similarity, and histogram similarity (HS) assesses color distribution. chamfer match score (CMS~\romannumeral1) focuses on average edge similarity, hausdorff distance (HD) emphasizes worst-case edge similarity, and contour match score (CMS~\romannumeral2) evaluates overall shape similarity.}
\label{tab:tab1}
\hspace{-1cm}
    \begin{tabular}{l||cccccc}
        \toprule
\textbf{Methods}    & \textbf{GM \(\downarrow\)}    & \textbf{HS \(\downarrow\)}     &\textbf{LPIPS \(\downarrow\)}   & \textbf{CMS ~\romannumeral1~\(\downarrow\)}    & \textbf{HD \(\downarrow\)}    & \textbf{CMS~\romannumeral2~\(\downarrow\)}\\
\midrule
Reference Only~\cite{zhang2023adding}           & 9.21e-06       & 0.89               & 0.71     & -0.13           & 212.25         & 5.76\\
Double ControlNet~\cite{zhang2023adding}           & 5.45e-06       & 0.81     & 0.60            & -0.10           & 209.88         & 8.38\\
Lora~\cite{hu2021lora}            & 3.43e-06       & 0.75                 & \textbf{0.52}            & -0.10           & 182.28         & 7.74\\
\hline
LPGen (Ours)            & \textbf{3.40e-06}       & \textbf{0.72}               & 0.55            & \textbf{-0.15}           & \textbf{154.54}         & \textbf{3.24}\\
\bottomrule
    \end{tabular}
\end{table*}
The landscape paintings generated by the proposed method were quantitatively compared with those generated by Reference Only, Double ControlNet, and Lora. Each model generated 15 images for each pair of style and structure, resulting in a total of 60 landscape painting images. For each model, the highest values for six metrics of the generated images were recorded: gram matrix similarity, histogram similarity, LPIPS, chamfer match score, hausdorff distance, and contour match score. The scores of the different diffusion models are shown in Table~\ref{tab:tab1}.

In Table~\ref{tab:tab1}, the first three metrics, the gram matrix, histogram similarity, and LPIPS, are used to analyze the structural similarity between model outputs and reference images. The table shows that LPGen performs best with respect to the gram matrix, suggesting that LPGen's generated images have the most similar texture style to the reference image. Furthermore, LPGen performs best with respect to histogram similarity, suggesting that LPGen-generated images have the most similar color distribution to the reference image. For structural similarity, LPGen excels in both gram matrix and histogram similarity, while Lora performs best in  LPIPS due to the structure controller. To repetitive fine-tuning, Lora performs best in terms of LPIPS, indicating that Lora's generated images are most perceptually similar to the reference image. Taking these factors into account, LPGen is the best model for overall structural Tsimilarity.

In Table~\ref{tab:tab1}, the last three metrics, the chamfer match score, hausdorff distance, and contour match score, are used to analyze the style similarity between different model outputs and the reference image, mainly focusing on edges and contours. Obviously, LPGen performs best with respect to the chamfer match score, suggesting that LPGen's generated images have the most similar edges to the reference image. Meanwhile, LPGen performs best with respect to the hausdorff distance, suggesting that LPGen's generated images are closest to the reference image regarding edge similarity. Furthermore, LPGen performs best with respect to the contour match score, indicating that LPGen's generated images have the most similar contour shapes to the reference image. Considered comprehensively, the landscape paintings generated by the proposed model have the most similar edges and contour shapes to the reference image, making it the best model for style similarity.

\subsection{Visual Assessment}

\begin{figure}
\centering
\includegraphics[width=1\linewidth]{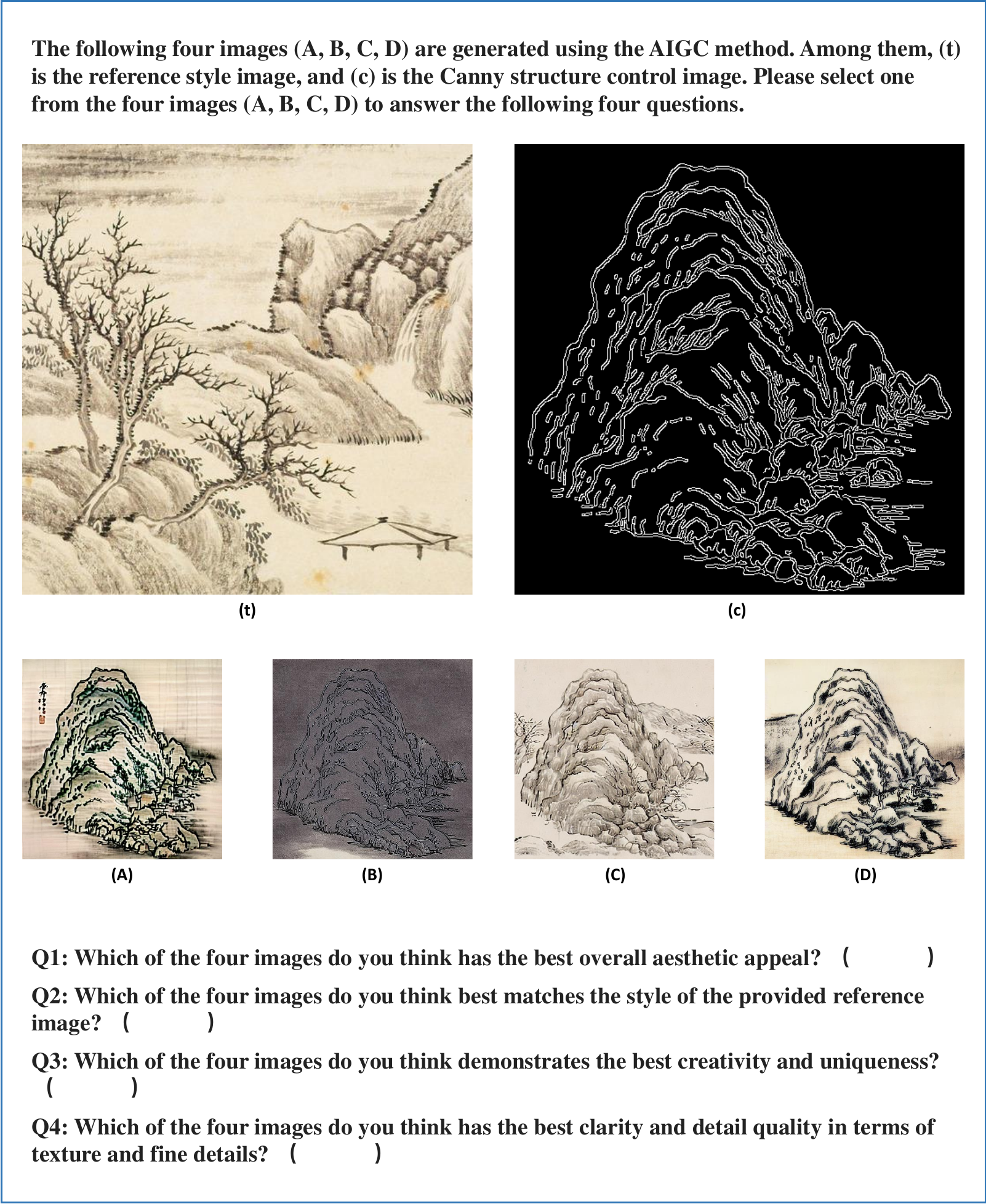}
\caption{Sample questions for a user survey. Each question presents four options, one of which is an image generated by this study. The questions evaluate the images' overall aesthetic appeal, style consistency, creativity, and detail quality.}
\label{fig:fig6_Question}
\end{figure} 

\begin{figure*}
\centering
\includegraphics[width=0.9\linewidth]{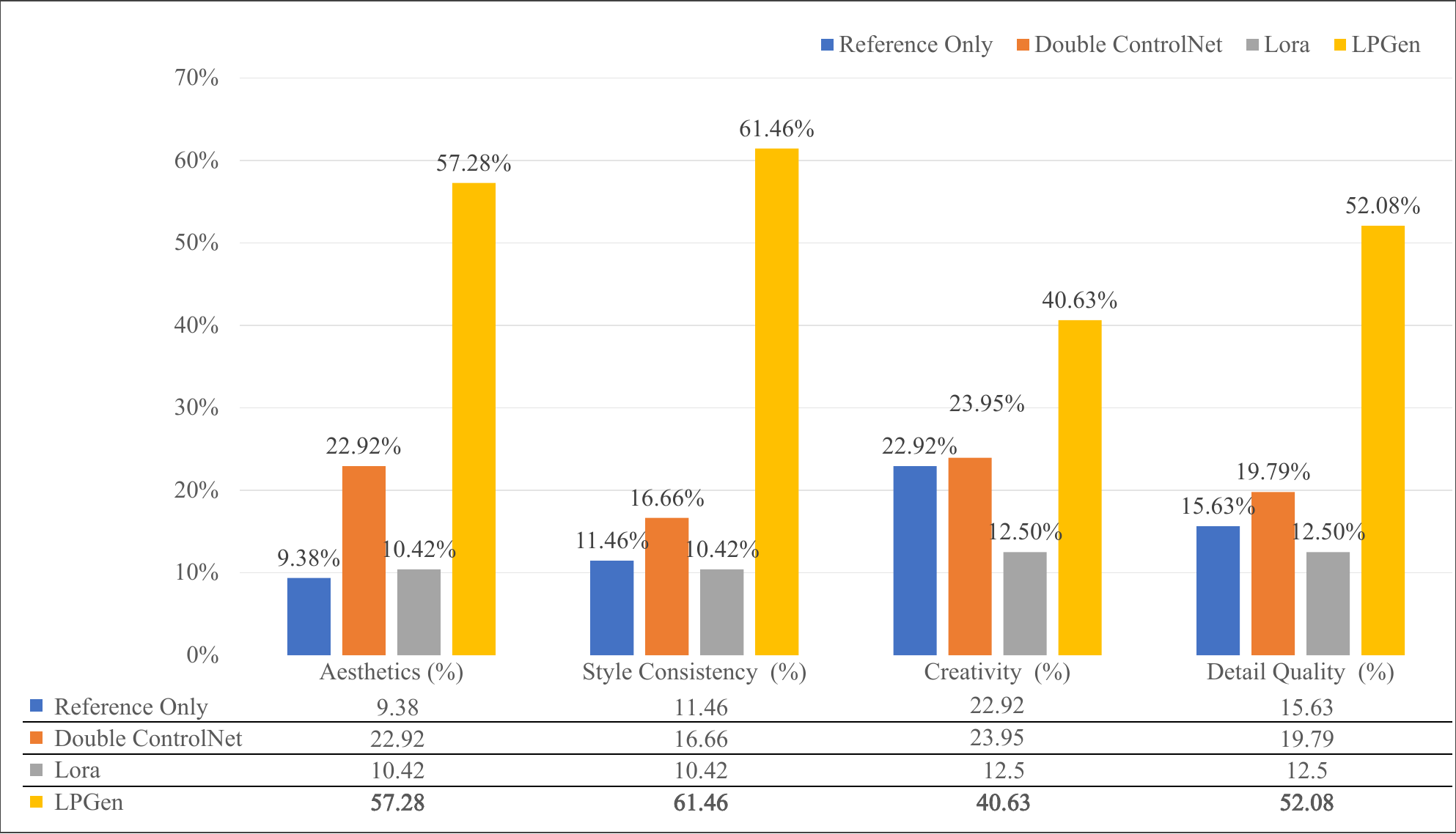}
\caption{Quantitative evaluation of the different models when generating landscape paintings. Each data represents the top-rated results, as determined by users, for images generated by different models in terms of aesthetic appeal, style consistency, creativity, and detail quality.}
\label{fig:fig7_Quantitative_evaluation}
\end{figure*} 
A total of 24 groups of landscape paintings were generated using different models: LPGen, Reference Only, Double ControlNet, and Lora, for a questionnaire. Additionally, 16 artists were invited to participate in the survey. In this research, we evaluated four critical aspects of the generated images: aesthetic appeal, style consistency, creativity, and detail quality. 
The aesthetic appeal metric evaluates the overall visual attractiveness of the images. The participants rated the images based on how pleasing they found them, considering factors such as color harmony, composition, and the emotional response evoked by the artwork.
The consistency aspect of the style examines how well the generated images adhere to a specific artistic style. Participants evaluated whether the images consistently incorporated stylistic elements of traditional landscape paintings, such as brushstroke techniques, use of space, and traditional motifs.
The creativity aspect measures the originality and innovation of the generated images. The participants rated the images based on the novelty and inventiveness of the compositions and interpretations within the confines of traditional landscape painting.
The detail quality metric focuses on the precision and clarity of the finer details within the images. Participants evaluated the quality of intricate elements such as textures, line work, and depiction of natural features such as trees, rocks, and water.

From Figure~\ref{fig:fig7_Quantitative_evaluation}, it can be seen that LPGen excelled with an impressive score of 61. 46\%, far exceeding the Reference Only model, the Double ControlNet model, and the Lora model in terms of style consistency. When it comes to evaluating creativity, LPGen once again led the pack with a top score of 40.63\%. In contrast, the Reference Only model scored 22.92\%, the Double ControlNet model 23.95\%, and the Lora model 12.50\%. Regarding detail quality, LPGen achieved the highest score of 52.08\%, showcasing its superiority in rendering intricate elements with precision and clarity. Moreover, LPGen has excellent aesthetic appearance, reaching the highest level of 57.28\%. Considering these results from all four metrics, the LPGen model's outstanding performance across four metrics—aesthetic appeal, style consistency, creativity, and detail quality—highlights its effectiveness in generating high-quality, artistically compelling landscape paintings. These results reflect the model's ability to meet and exceed user expectations in various aspects of image generation.

\subsection{Generated Showcase}

\begin{figure*}
\centering
\includegraphics[width=13.8cm]{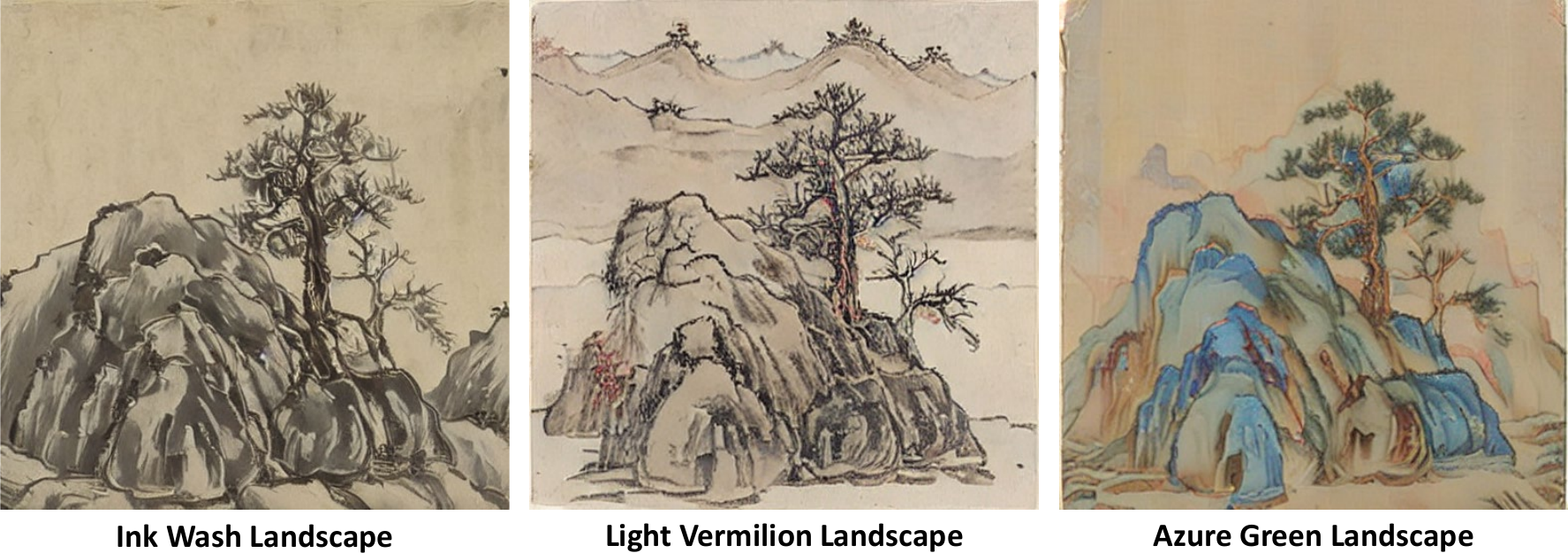}
\caption{Landscape paintings generated by the proposed method using the same structure reference image but different style images. The generated images have distinct stylistic features such as azure green landscape, ink wash landscape, and light vermilion landscape}
\label{fig:fig8_Showcase}
\end{figure*} 

Figure~\ref{fig:fig8_Showcase} shows the landscape paintings generated by the proposed method, which is in the style of the famous such as azure green landscape, ink wash landscape, and light vermilion landscape. The landscape paintings generated met the style requirements and were full of details. Specifically, the techniques in landscape painting such as outlining, chapping, rubbing, moss-dotting, and coloring are clearly visible, with distinct stylistic features that accurately capture the essence of traditional Chinese landscape painting. In addition, the three images initially used the same structure reference image. The generated images had a good composition, accurately distinguishing mountains, trees, and distant views. Therefore, the architectural design generated by the proposed method reached a usable level, capable of reducing creative difficulty and improving creative efficiency.  

\section{Discussion}
This study demonstrates the effectiveness of the proposed method through both qualitative and quantitative analyses. In terms of qualitative analysis, visual comparisons with other methods show that the proposed method can generate landscape paintings with specified styles and structures, a capability that mainstream models lack. On the quantitative side, the data prove the superiority of the proposed method across all evaluation metrics. In particular, the method excels at generating specified styles and structures. For example, in the histogram similarity metric, the proposed method performs well, with values of 0.89, 0.81, and 0.75, respectively, lower than those of reference only, double controlNet, and Lora, thus convincingly demonstrating its effectiveness.

By directly utilizing AI to generate diverse landscape paintings, LPGen replaces the tedious processes of outlining, chapping, rubbing, moss-dotting, and coloring in the traditional hand-drawing method. Compared to conventional methods, LPGen excels in both efficiency and creative generation. Regarding design efficiency, traditional methods typically require approximately two days to complete a creation and its corresponding revisions, whereas LPGen, running on a consumer-grade GPU with 8GB VRAM, can generate a complex image description using Stable Diffusion in around 4 seconds. As computational power continues to improve, the speed of generating interior design videos with LPGen can be further enhanced. In terms of creative design, LPGen offers style options for users to choose from, allowing for either hand-drawn structure diagrams or automatic extraction of structure diagrams from other images, thereby accelerating the design process. In general, LPGen demonstrates the feasibility of innovative methods for generating landscape images. Additionally, LPGen is highly scalable; by replacing the underlying diffusion model, it can be adapted to other generation painting.

The content generated by AI will profoundly impact current painting methods. In terms of efficiency, AI will increasingly take over tasks that emphasize logical and rational descriptions, ultimately forming an AI design chain. Simple design tasks will be completed by AI, thereby giving designers more time to focus on creativity and improve quality. In terms of role positioning, designers are no longer just traditional creators; they are transitioning to facilitators collaborating with AI. For example, in this study, the artist's role is not merely to draw images; they use their expertise to collect and organize data and transfer knowledge to the AI model. This new human- machine collaboration approach is likely to become the norm in future digital painting, driving the design process toward automation and intelligence. 

\section{Conclusions}
This paper presents LPGen, a novel diffusion-based model that addresses the challenge of generating high-fidelity landscape paintings with a balanced control over structure and style. By introducing a decoupled cross-attention mechanism, LPGen effectively processes structural and stylistic features independently, reflecting the layered techniques used in traditional painting. The integration of a structural controller further enhances the model’s ability to maintain aesthetically pleasing compositions. Pre-trained on a curated dataset of high-resolution landscape images and fine-tuned for detailed output, LPGen consistently outperforms existing models in both structural accuracy and stylistic coherence. This work not only advances the field of AI-generated art but also bridges technology with traditional artistic practices, offering valuable insights for future research. The public release of our code, dataset, and model weights will enable broader exploration and application of LPGen in creative domains.

\vspace{-9.5cm}
\begin{IEEEbiography}[{\includegraphics[width=1in,height=1.25in,clip,keepaspectratio]{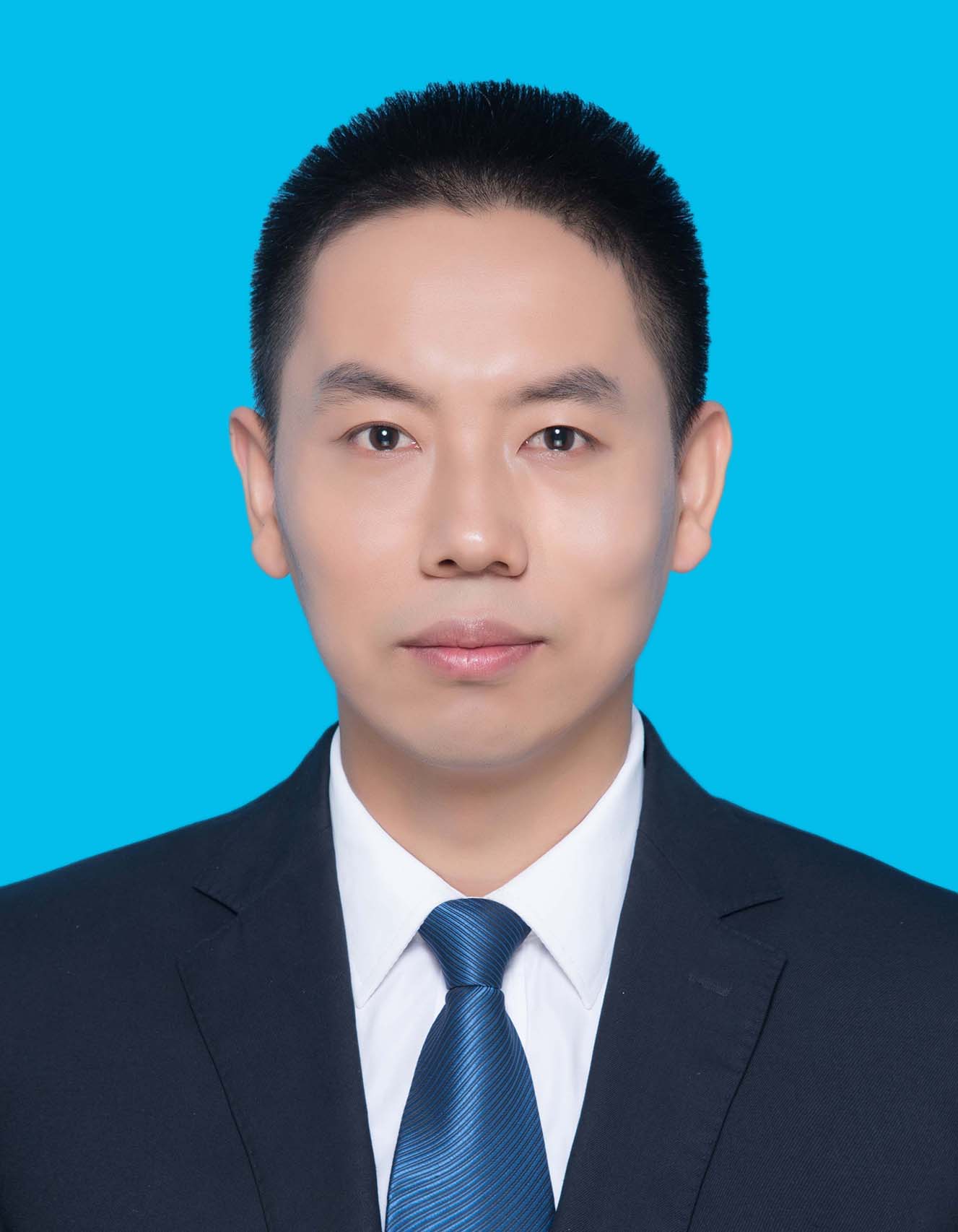}}]{WANGGONG YANG} received the B.Sc. and M.Sc. degrees in communication engineering from Chongqing University of Posts and Telecommunications, China, in 2005 and 2008. Since 2009, he has been with the School of New Media, Beijing Institute of Graphic Communication, China, where he is currently an Associate Professo. His research interests include digital media art, artificial intelligence art design, virtual reality and interactive design.
\end{IEEEbiography}
\vspace{-9.0cm}

\begin{IEEEbiography}[{\includegraphics[width=1in,height=1.25in,clip,keepaspectratio]{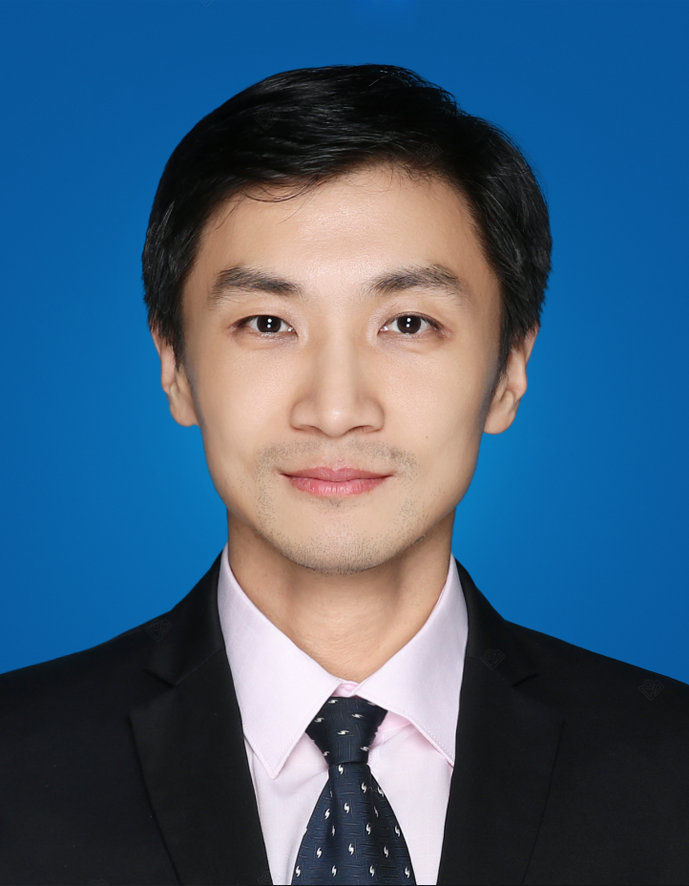}}]{Zhao Yifei} received the Master's degree in Engineering from Beijing Institute of Technology in 2012. From July 2004 to June 2016, worked as a teaching assistant and lecturer at the School of  Art \& Design, Beijing Institute of Graphic Communication. From July 2016 to present, he has served as an associate professor and master's supervisor at the School of New Media at Beijing Institute of Graphic Communication, as well as assistant dean and head of the Department of Digital Media Arts. His main research directions include digital media art, virtual reality art design, and artificial intelligence assisted art design.
\end{IEEEbiography}

\newpage

\EOD

\end{document}